\documentclass[conference]{IEEEtran}
\IEEEoverridecommandlockouts
\usepackage{cite}

\usepackage{hyperref}
\usepackage{commath}
\usepackage{amsmath,amssymb,amsfonts, array}
\usepackage{chemformula}
\usepackage{caption}
\usepackage{algorithm}
\usepackage{algpseudocode}
\usepackage{subcaption}
\usepackage{textcomp}
\usepackage{xcolor}
\usepackage{array}
\usepackage{stackengine}
\usepackage{euscript}[mathcal]
\usepackage[inline]{enumitem}
\usepackage{multirow}
\usepackage{amsmath}
\usepackage{amsthm}
\usepackage{booktabs}

\def\BibTeX{{\rm B\kern-.05em{\sc i\kern-.025em b}\kern-.08em
    T\kern-.1667em\lower.7ex\hbox{E}\kern-.125emX}}
\begin{document}
\newcommand\barbelow[1]{\stackunder[1.2pt]{$#1$}{\rule{.85ex}{.085ex}}}

\title{
Negative Selection Approach to support Formal Verification and Validation of BlackBox Models' Input Constraints
}

\author{ Abdul-Rauf Nuhu$^{1}$, Kishor Datta Gupta$^2$, Wendwosen Bellete Bedada$^1$, Mahmoud Nabil$^1$, Lydia Asrat Zeleke$^1$,\\ Abdollah Homaifar$^{1,*}$, and Edward Tunstel$^3$
\thanks{$^1$ North Carolina A\&T State University, Greensboro, North Carolina, US, 27411, Emails: {\tt\small anuhu@aggies.ncat.edu, wbbedada@ncat.edu,  mnmahmoud@ncat.edu, lzeleke@aggies.ncat.edu, homaifar@ncat.edu}}
\thanks{$^2$ Clark Atlanta University, Atlanta, GA USA. Email:
{\tt\small kgupta@cau.edu}.}
\thanks{$^3$ Motiv Space Systems, Inc., Pasadena, CA, 91107 USA, Email: {\tt\small tunstel@ieee.org}}
\thanks {$^*$ Corresponding Author}
}

\maketitle

\begin{abstract}
Generating unsafe sub-requirements from a partitioned input space to support verification-guided test cases for formal verification of black-box models is a challenging problem for researchers. The size of the search space makes exhaustive search computationally impractical. This paper investigates a meta-heuristic approach to search for unsafe candidate sub-requirements in partitioned input space. We present a Negative Selection Algorithm (NSA) for identifying the candidates' unsafe regions within given safety properties. The Meta-heuristic capability of the NSA algorithm made it possible to estimate vast unsafe regions while validating a subset of these regions. We utilize a parallel execution of partitioned input space to produce safe areas. The NSA based on the prior knowledge of the safe regions is used to identify candidate unsafe region areas and the Marabou framework is then used to validate the NSA results. Our preliminary experimentation and evaluation show that the procedure finds candidate unsafe sub-requirements when validated with the Marabou framework with high precision.

\end{abstract}


\section{Introduction}
Machine learning algorithms have revolutionized the way complex systems are designed and evaluated by leveraging the accessibility of data. In particular, the usage of Deep Neural Networks (DNNs) in safety critical domains is growing increasingly, spanning areas such as autonomous systems \cite{bloom2017self}, avionics industries \cite{conner2015nasa} and GPS attack detection \cite{agyapong2021efficient}. Data-driven models are also used in the transportation sector for traffic volume prediction \cite{lartey2021xgboost}, driver behavior identification \cite{girma2019driver} and in multi-UAV environments for testing \& evaluating UAV behavior \cite{sarkar2021framework}. Data-driven models in these domains have shown some impressive results including matching if not surpassing the performance of manually programmed software \cite{hou2013model}. 

The success of data-driven models necessitates developing and implementing formal analysis and verification frameworks for verification and certification purposes.
A DNN verification process primarily consists of two components \cite{katz2019marabou}, (i) a trained neural network $N$ and (ii) a property $P$ to be checked. The verification process addresses whether the safety property is satisfied or if it returns a concrete example that violates the given property \cite{katz2019marabou}. Most verification frameworks address reachability  and adversarial robustness aspects \cite{katz2019marabou,tran2020nnv}. Reachability analysis involves defining constraints, or using a set representation \cite{tran2020nnv} which encapsulates the entire input domain as safety constraints, and propagates these constraints through the layers of the DNNs to compute an estimated output. The estimated output is compared to the given output safety constraints and conclusions can be drawn.

However, a recently growing aspect of verification is verification-guided test case generation which aims to improve the performance of data-driven models by utilizing counterexamples extracted during the verification. Verification-guided test case generation as proposed in \cite{sharma2020higher},\cite{sharma2021mlcheck}  uses the counterexamples of the verification frameworks to retrain the models. However, verification techniques in many cases return only one instance in the case where a requirement is violated. Using a single instance is not sufficient in validating and testing a trained machine learning model. Recent approaches \cite{katz2019marabou, wang2018formal,wu2020parallelization, gopinath2018deepsafe} suggest finding unsafe regions from partitioned input space to guide the generation of multiple counterexamples at a time. Determining the unsafe regions within the partitioned input space involves verifying all of the sub-requirements; and with large partition factor or number of inputs, they can result in overwhelmingly large sub-input domains. Our literature survey revealed that use of high-dimensional datasets based on DNN makes it hard to verify and validate the DNN utilizing input sub-constraints, and an exhaustive search is not computationally feasible. Therefore, an intelligent searching approach that governs finding critical sub-input domains (input sub-requirements highly vulnerable to violating the safety requirement) is needed to prune the sub-input domain. 

 The advantage of utilizing a heuristic approach is that it offers a fast, easy-to-understand solution implementation. Heuristic algorithms are practical, serving as quick and achievable short-term solutions to planning and scheduling problems. Meta-heuristic algorithms provide randomness that makes it possible to generate different unsafe regions for each test session, which makes it more practical for the formal-verification problem.
The Negative Selection Algorithm (NSA) is one of the leading meta-heuristic algorithms\cite{9546626} and is being applied in safety critical areas such as anomaly detection \cite{ji2007revisiting} and computer security problems \cite{forrest1994self}. However, the NSA has not been applied to guide the search of candidate unsafe sub-requirements, and we leverage it for finding the critical (unsafe) regions within a given violated safety requirement. Our work proposes verification within small input domains by partitioning the original space to help find the safe and unsafe regions within the input space of a given requirement.\\ 

The contributions of the paper comprises the following:

\begin{itemize}
    \item A simple yet effective input-based partition strategy for generating sub-requirements from a given safety requirement is presented.  
    \item  NSA (based on the prior knowledge of the verification status of the sub-requirements) is applied to do a meta-heuristic search for unsafe sub-input domains within large sub-input domains.
    \item  Experimental demonstration of the proposed method is shown with the Marabou framework.
\end{itemize}
\par Section \ref{NSA} presents background on the underlying theory, approaches, and preliminaries covering the problem formulation. This is followed by presentation of the proposed methodology in Section \ref{Proposed-method}. Section \ref{results} presents the experimental results and Section \ref{Conclusion} concludes and outlines the future directions.
\section{Preliminaries}\label{NSA}
In this section, we highlight the background on Splitting Strategies, approaches based on the satisfiability modulus theorem (SMT), and the NSA approach followed by a brief description of our problem formulation.

\subsection{Splitting input space of a given safety requirement}
Smaller input sub-requirements help in finding safe and unsafe regions within a violated safety requirement. Additionally, they can allow most non-linear activation functions to exhibit linear behavior \cite{gopinath2018deepsafe}. Given a safety requirement described using the following parameters;
\\
\\ $d$ input Dimensions:  $\forall i \in \{1,..,d\}$
\\ Lower and upper input bounds: $ l \leq X_i\leq u $, where $ l, u \in \mathbb{R}$
\\
\\
the splitting approach involves partitioning the ranges of the input variables of the original requirement into sub-intervals \cite{katz2019marabou}. The input variable $X_{i}$ is split into $K$ continuous non-overlapping sub-intervals. Consider a given input requirement on a network as: {$\alpha := \alpha' \wedge ( -1 \leq X_{1} \leq 1 ) \wedge ( -2 \leq X_{2} \leq 1 )$,}\\ where $X_{1}$ and $X_{2}$ are the two input variables encoded by $\alpha'$ \cite{wu2020parallelization}. Fig. \ref{splitting} illustrates the splitting of input variable $X_1$ of the given safety requirement into $K=2$ sub-intervals.

\subsection{SMT-based approaches}
SMT-based formal verification frameworks reduce a data-driven model verification problem to a constraint satisfiability problem. SMT-based approaches used custom satisfiability (SAT) solvers and are noted for being sound and complete. Existing SMT-based frameworks are designed for DNN, a class of data-driven models. These approaches encode a given DNN model, given input safety requirements and the negation of the desired output safety requirement as set of constraints. The encoded constraints are given to a SAT solver, if it finds a \textbf{sat} (satisfiable) assignment on the constraints, the corresponding assignment solution is a counterexample to the safety requirement/property. However, if the SAT solver returns \textbf{unsat} (unsatisfiable) on the constraints, it means no counterexample exists and the safety requirement/property holds. For instance, consider an $n$ layer, single-output feed-forward and fully connected neural network with rectified linear unit (ReLU) functions after each hidden layer. Further, consider a safety requirement specifying a bounded input domain and requiring the output value of the network to be a positive value. Formally verifying such a safety requirement can be framed as the following SMT problem \cite{bunel2018unified}:
\begin{subequations}\label{subeqns}
\begin{align}
{\barbelow{X}_i} & \leq X_i \leq \bar{X}_i \hspace{2.2cm} \forall i \in \{1,..,d\}\label{1_a}\\
\widehat{X}_{j+1} & = W_{j+1}X_j + b_{j+1} \hspace{1.1cm} \forall j \in \{1,..,n-1\}\label{1_b}\\
X_{j} & = max\{0, \widehat{X}_{j}\}  \hspace{1.6cm} \forall j \in \{2,..,n-1\}\label{1_c}\\
X_{n}&\leq 0 \label{1_d}
\end{align}
\begin{figure}[h]
\centering
\includegraphics[width=0.65\linewidth]{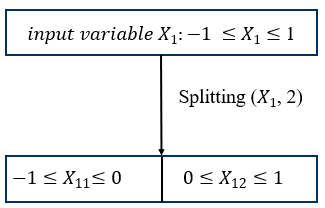}
\caption{Splitting an input variable of a safety requirement into $2$ sub-intervals.}
\label{splitting}
\end{figure}
 \begin{figure}[]
\centering
\includegraphics[width=1.0\linewidth]{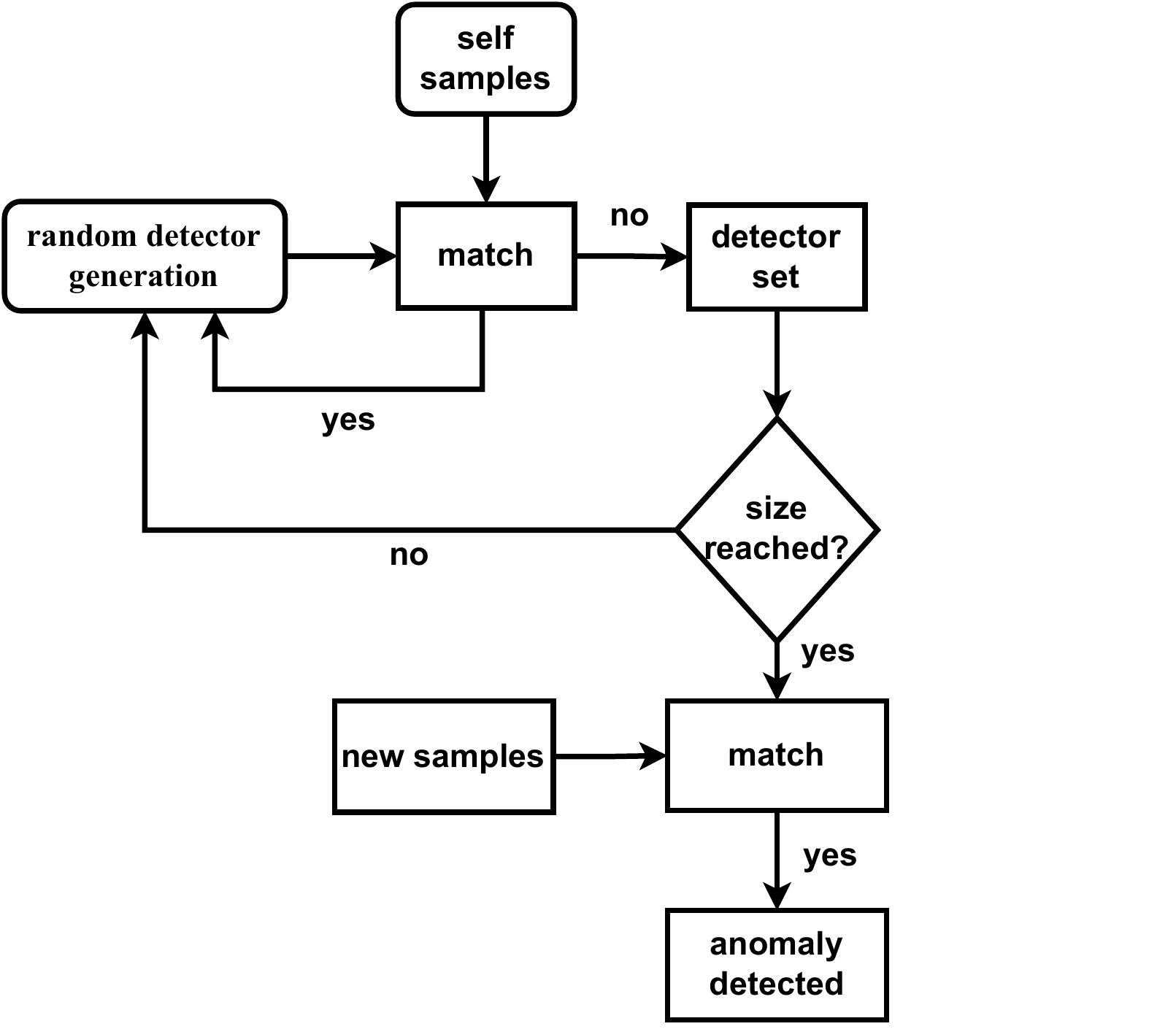}
\caption{The Negative Selection Algorithm (NSA) procedure, consist of two phases: training and testing phase, training phase generates detector set and the testing phase classifies new samples whether they belong to defect or clean class \cite{9546626}. }
\label{fig:bas}
\end{figure}
\end{subequations}

Equations (\ref{1_a}) and (\ref{1_d}) respectively represent the constraints on the inputs and the negation of the constraints on the neural network's output. Equation (\ref{1_b}) and (\ref{1_c}) respectively represent the encoding of the affine transformations performed by the hidden layers and ReLU activation functions. A value assignment to the problem variables that satisfies all the constraints gives a valid counterexample \cite{katz2017reluplex}. However, if there is no value assignment to the problem variables that are satisfied (unsatisfiable assignment), the safety requirement is verified to be safe. The safety requirement and trained DNN model are encoded using the encoding procedure in \ref{subeqns}. Then the framework finds a value assignment to the problem variables that satisfies all the constraints; if yes, then that instance corresponds to a counterexample, otherwise, the given safety requirements are valid and verified.\\
\indent The Marabou framework \cite{katz2019marabou} is an SMT-based approach built on the Reluplex framework \cite{katz2017reluplex}, which uses and extends on the Simplex algorithm $-$ a standard decision procedure for conjunctions of linear constraints, to verify DNN models. The framework encodes the input bounds, output bounds, weighted sums for the hidden layers as linear constraints and activation functions as non-linear constraints. It iteratively searches for an assignment that satisfies all the constraints on the given requirements, while treating the non-linear constraints lazily hoping that most of them will prove irrelevant to the requirement under consideration.
\subsection{Negative selection Algorithm (NSA)} 
NSA is a bio-inspired method and a robust computational mechanism for classification problems where the decision exists in the complementary space of positive space\cite{9546626}. An NSA consists of two phases: the training and detection phase. The training phase generates the detector set while the detection phase examines new samples using the detector set. Different types of NSA exist but in this work we focus on R-NSA (real valued NSA). R-NSA generally works as one-class classification, it utilizes positive data (self) to create negative data areas (detectors) in the training time, and at the testing time it matches with the data area to identify whether the data is positive or negative. As it utilizes real value representation, it is easier for R-NSA to work with continuous data and it can take any radius size for detectors. In Figure \ref{fig:bas}, the basic flow of the R-NSA algorithm is presented. At first it randomly generate some data points and checks whether these data are self or not using a matching rule (Distance measure). The random data sample distance is measured from the self points, if these data are not self then they generate an area around the data point based on radius size and considers these as non-self regions called detectors. Depending on the data representation, similarity measures such as Euclidean distance, Manhattan distance, R-bit chunk matching, Hamming distance, etc., can be used. 

\subsection{Problem Formulation}
Let $\mathbf{P}=[p_1,p_2,\dots,p_Z]$ be the partitioned input space comprised of sub-requirements and $\mathbf{H}$ a heuristic or meta-heuristic algorithm which takes $\mathbf{P}$ as input and generates a set of candidate unsafe sub-requirements, $\mathbf{D}$. Thus, $\mathbf{H}:\{\mathbf{P}\}\longrightarrow\mathbf{D}$. Also, there exists a verification framework, $\mathbf{V}$ which takes $\mathbf{D}$ as input and outputs \textbf{sat} or \textbf{unsat} on each of $\mathbf{D}$.\\
We decomposed the problem into three sub-problems as listed below:
\begin{enumerate}
    \item[] \textbf{Sub-problem 1:} How can we generate the partitioned input space, $\mathbf{P}$, comprised of sub-requirements?
    \item[] \textbf{Sub-problem 2:} Having $\mathbf{P}$, how can we generate input regions of interest, $\mathbf{D}$, using heuristic or meta-heuristic algorithms?
    \item[] \textbf{Sub-problem 3:} Having $\mathbf{D}$, how can we determine the number of $\mathbf{D}$ members that are unsafe?
\end{enumerate}
\section{Proposed Approach}\label{Proposed-method}
\begin{figure}[h]
    \centering
    \includegraphics[width=1\linewidth]{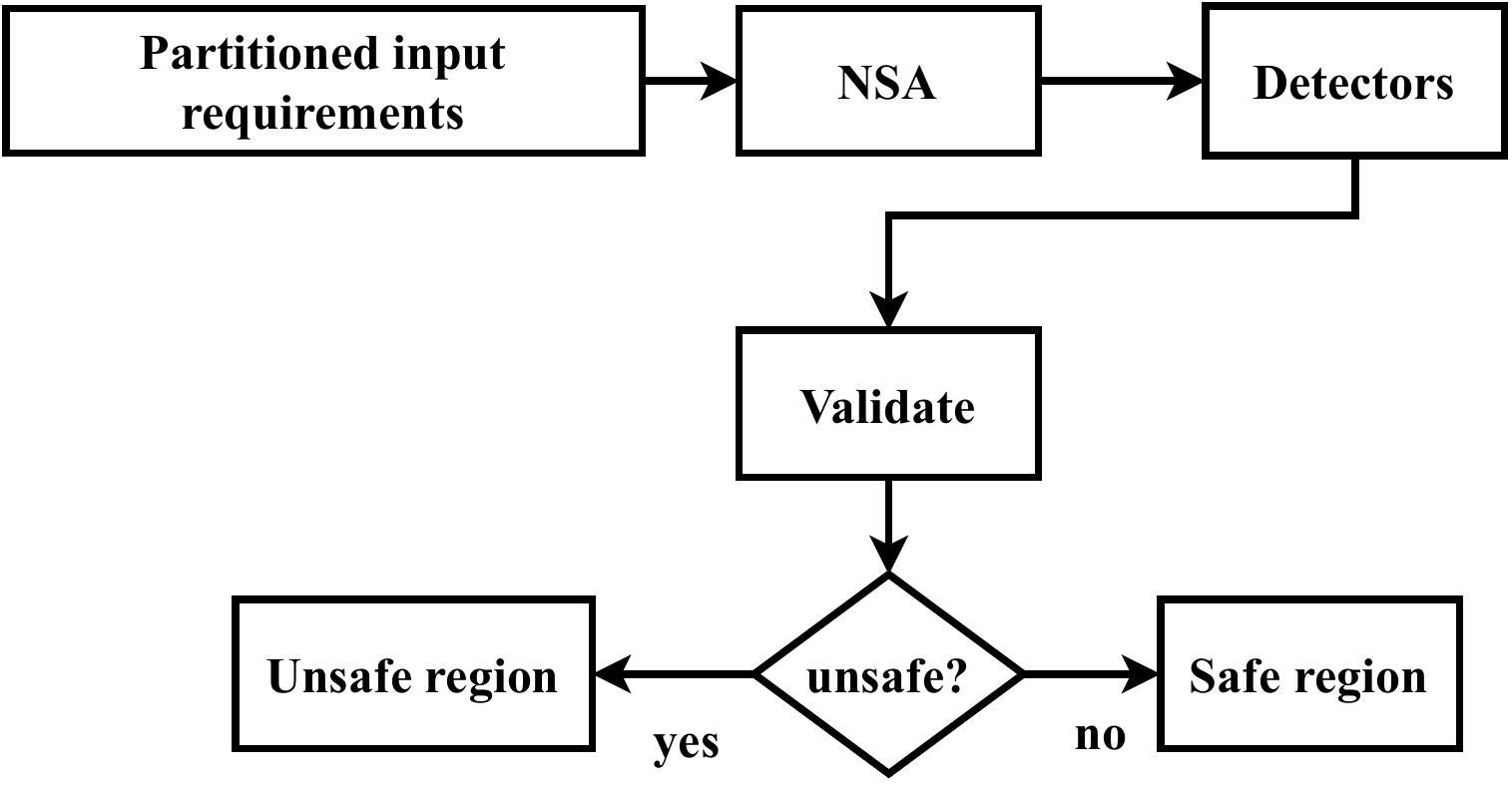}
    \caption{Overview and Workflow of the proposed framework. It takes as input the partitioned input space, then the NSA procedure finds candidate unsafe members (Detectors); a verification framework is then used to validate whether the Detectors are unsafe or not. }
    \label{proposed-framework}
\end{figure}

The workflow of the proposed approach is presented in Fig. \ref{proposed-framework}. It takes the partitioned input sub-requirements as input. Candidate unsafe sub-requirements are generated from the partitioned input sub-requirements by applying the NSA. We then validate these sub-requirements to assess the effectiveness of the NSA. In summary, the proposed approach consists of two main components: the generation of the input sub-requirement and the NSA implementation.
\begin{figure}[h]
\centering
\includegraphics[width=0.5\linewidth]{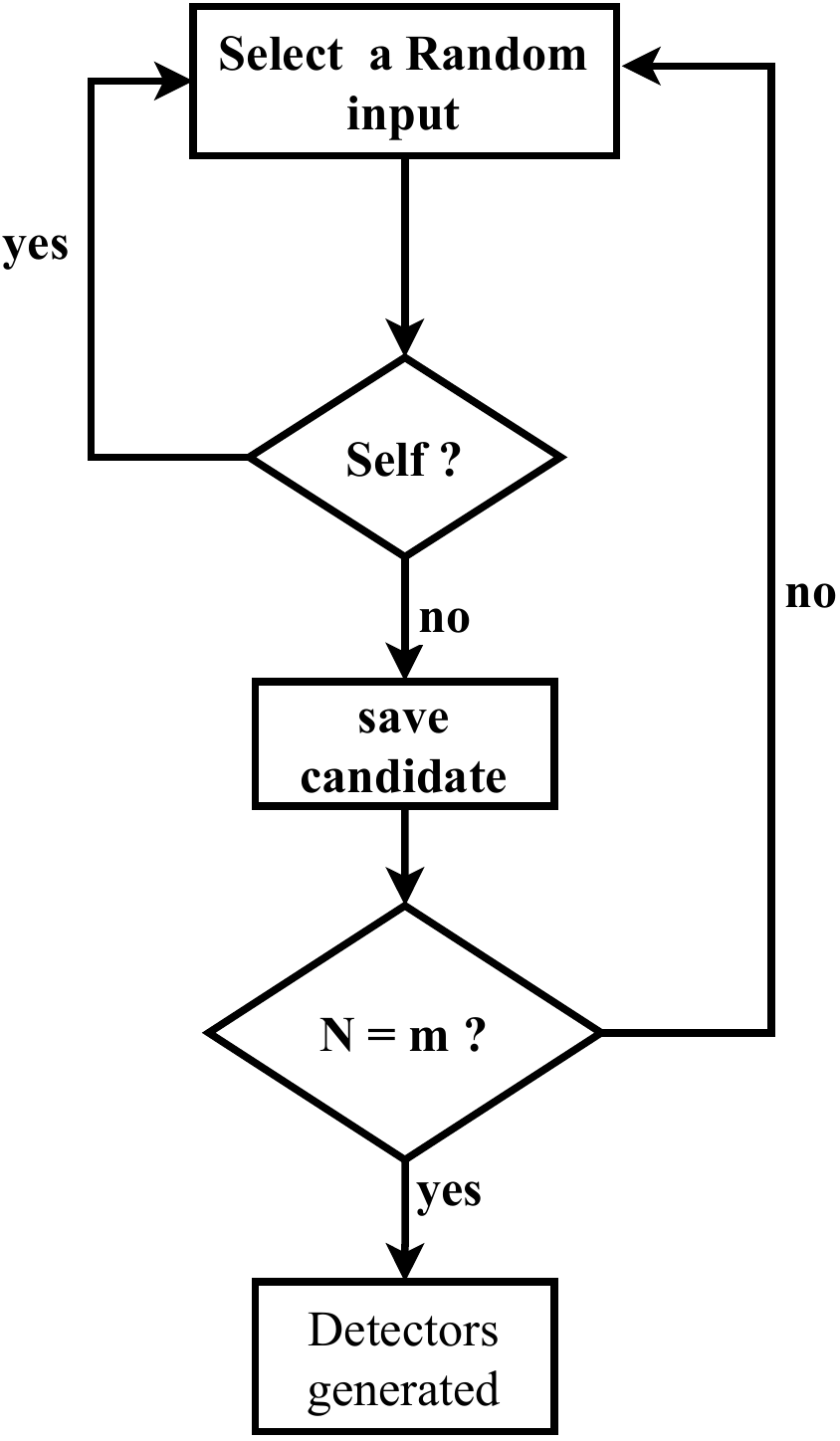}
\caption{NSA detector generation selects an element randomly from an input space  and compares it with the safe members (self). Members dissimilar from the self are stored as detectors (candidate unsafe members).}
\label{nsa-approach}
\end{figure}

\begin{algorithm}[tb]
\footnotesize
   \caption{Detectors generation procedure}
   \label{alg:example}
\begin{algorithmic}
   \State {\bfseries Input:} \textbf{P}: Partitioned input requirements, 
    \textbf{S}: Set of self samples,
    \textbf{$r_{s}$}: Self radius, 
    \textbf{N}: detectors threshold\\
   \State {\bfseries Output:} \textbf{Detectors}: D
   \State {$m \leftarrow 0 $}
   \State {$X \leftarrow \emptyset$}
   \While{$ m $ $< N$}
   \State $x \leftarrow $ uniformly select random sample from \textbf{P} ;
   \If{$x \notin X$}
   \For {$y$ in \textbf{S}}
   \State $D \leftarrow \emptyset$
   \If {$||x -y|| > r_{s}$}
   \State {D = D $\cup$ \{x\}}
   \State $m \leftarrow m + 1$
   \EndIf
  \EndFor
  \State{$X = X \cup x$}
  \EndIf
  \EndWhile 
  \State return D 
\end{algorithmic}
\end{algorithm}

\subsection{Input sub-requirements generation} 
In our approach, we take a set of inputs from the sub-input regions and check if they are similar or not to the safe regions based on a radius value. Ground truth safe regions from a verification framework are needed, and the verification frameworks in \cite{katz2019marabou} can guide the generation of the ground truth safe input sub-requirements. However, partitioning approaches in this work are designed for improving execution and handling over-approximation challenges. Therefore, we proposed a simple yet effective partitioning strategy to generate sub-intervals based on which sub-requirements are generated. Our partition/splitting approach starts by first generating sub-intervals for all input variables with significantly large ranges. Assume $i \in \{1,..,d\}$ and $j \in \{1,..,n\}$, where $d$ is the dimension of the input variables considered in the input space and $n$ is the user-specified number of sub-intervals, and assume bounds for each dimension as ${\barbelow{X}_i}\leq X_i\leq \bar{X}_i$. The partitioning step size for each input dimension can be computed by Equation \ref{sub:range} as
\begin{equation}
    \gamma_{i} = \frac{{\bar{X}_i}-{\barbelow{X}_i}}{n}
    \label{sub:range}
\end{equation}
Once the partitioning step size is given, the sub-intervals are computed for the $i^{th}$ dimension and $j^{th}$ element by utilizing Equation \ref{sub:interval}.
\begin{equation}
    X_{ij} \subseteq \mathbb{R} \quad | \quad [{\barbelow{X}_i}+[j-1]\gamma_{i}\leq X_{ij} \leq {\barbelow{X}_i}+[j]\gamma_{i} ]
    \label{sub:interval}
\end{equation}
The set of all sub-intervals for the $i^{th}$ dimension is given as 
\begin{equation}
    X_i = \{X_{i,1},\hdots, X_{i,n} \}.
\end{equation}
Once the sub-intervals are obtained, we combine sub-intervals from each dimension to form sub-requirements using a Cartesian product. Letting $P$ denote a list of sub-requirements, 
\begin{equation}
    P = \{P_1,\hdots,P_Z\}, \textrm{where} \quad Z = n^d 
\end{equation}
each sub-requirement, $P_z$, is given as an element of the cross-product of sub-intervals 
\begin{equation}
    P_z \in \{X_1 \times X_2, \hdots, \times X_d\}.
\end{equation}
\subsection{NSA procedure}
Our NSA procedure for obtaining candidate unsafe sub-requirements (detectors) consists of the training phase of the NSA in Sec. \ref{NSA}$\--$ first, an input from the partitioned input space is randomly selected and compared with the ground truth safe input sub-requirements (self). Input sub-regions dissimilar to the safe sub-requirements based on the $L_1$ norm are stored as the detectors. Fig. \ref{nsa-approach} summarizes the detector generation procedure during the training phase of the NSA. Algorithm \ref{alg:example} shows the implementation of detector generation. First, we initiate a counter $m$ which keeps track of the generated detectors, create an empty set $X$ which ensures no individual repetition in the detectors. We randomly select individual $x$ from the partitioned space and compare it with individual $y$ from the self set. If the two individuals are dissimilar based the radius value $r_s$, we save individual $x$ as part of the detectors, $D$. Once the number of detectors, $m$ equals a user defined threshold value, $N$ the NSA procedure terminates and the detectors are stored as the candidate unsafe sub-requirements for validation using a verification framework.
\section{Experiments and Results} \label{results}
In the following, we discuss our implementation of the proposed approach and evaluate its performance on partitioned input safety requirements of a DNN-based controller.
\subsection{Test Case}
\begin{figure}[!htp]
\centering
\includegraphics[width=0.45\textwidth]{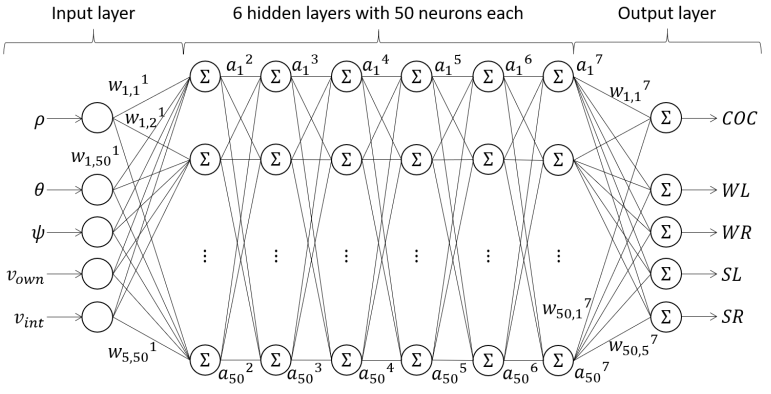}
\caption{The ACAS Xu DNN structure comprises $5$ inputs and outputs respectively and $6$ hidden layers each with $50$ nodes \cite{manzanas2021verification}.}
\label{dnnStructure}
\end{figure}

The DNN prototype of the unmanned variant of Airborne Collision Avoidance System X (ACAS Xu) is used as a test case for evaluating the proposed approach. ACAS Xu uses a lookup table in the avionics to issue horizontal maneuver advisories to an aircraft under consideration (\emph{ownship}) to avoid near midair collision with nearby aircraft (\emph{intruders}) \cite{katz2017reluplex}. The original ACAS Xu lookup table has seven inputs that represent information read from sensor measurements \cite{katz2017reluplex}:
(i) $\rho$: Distance from ownship to intruder, (ii) $\theta$: Angle to intruder relative to ownship heading direction, (iii) $\psi$: Heading angle of intruder relative to ownship heading direction, (iv) $v_{own}$: Speed of ownship, (v) $v_{in}$: Speed of intruder, (vi) $\tau$: Time until loss of vertical separation and (vii) $a_{prev}$: Previous advisories. There are five outputs which represent the different horizontal advisories that can be given to the ownship: Clear-of-Conflict (COC), weak right (WR), strong right (SR), weak left (WL), or strong left (SL). Due to memory constraints DNN were used to compress the original lookup table. $45$ DNNs were derived and each denoted as $N_{xy}$, where $x$ corresponds to the index of the previous advisory  $a_{prev}$: [COC, WL, WR, SL, SR ] and $y$ corresponds to the index of the discretized $\tau: [0, 1, 5, 10, 20, 40, 60, 80, 100 ]$. For instance, $N_{21}$ corresponds to a network with $a_{prev}$ = WL and $\tau$ = 0.  Therefore, each network takes the remaining five state variables as inputs and outputs a value associated with each of the five output variables $a_{prev}$: [COC, WL, WR, SL, SR ] \cite{katz2017reluplex}. Fig.\ref{dnnStructure} shows the network structure, each network is feedforward fully connected with ReLU activation function, five input and output nodes respectively, $6$ hidden layers with $50$ nodes each.  WL or WR means heading left or right, with an angular rate of 1.5$^0$/s. SL or SR denotes heading left or right with an angular rate of 3.0$^0$/s.

\subsection{Experimental setup}\label{sub-requirement}
The DNN prototype of the unmanned variant of Airborne Collision Avoidance System X (ACAS Xu) has $10$ properties comprising constraints on the inputs and outputs variables of the networks \cite{katz2017reluplex}. Among the 10 properties specified in the Appendix of \cite{katz2017reluplex}, property $\phi_2$ is considered because it contains both \textbf{sat} \& \textbf{unsat} verification outcomes on $36$ ACAS Xu DNN prototypes $\--$ $34$ networks with \textbf{sat} verification outcome and $2$ networks with \textbf{unsat} verification outcomes. Our main goal is to use the NSA to search candidate unsafe input sub-requirements. So, properties with verification status \textbf{sat} on the networks are essential for the proof-of-concept of the NSA approach. The property is normalized as required by the verification framework. To generate the sub-intervals, we considered input dimensions with considerable large range between the lower and upper bounds. Input dimensions having lower and upper bounds that were almost the same were kept the same throughout all the generated sub-requirements. Three input dimensions [$\rho$, $\theta$, $\psi$] were considered because the other two input dimensions ($v_{own}$, $v_{in}$) are almost constant in the normalization process.
The user-specified number of sub-intervals, ($n$) is set to $4$ since, empirically this results in less overlap in the sub-requirements bounds. We generated, in all, $64$ ground truth input sub-requirements\footnote{ Sub-requirements total number = n$^d$} from the sub-intervals using the Cartesian product. Table \ref{table-origRequirement} shows the normalized input constraints of property $2$, $\phi_2$ for the considered input variables for splitting. Table \ref{table-subrequirements} shows some of the generated input sub-requirements.
\begin{table}[htbp]
\scriptsize
\begin{center}
\begin{tabular}{ |c|c|c|c| } 
  \hline
  $\rho$ bounds & $\theta$ bounds & $\psi$ bounds \\ 
  \hline
  $0.6\leq \rho \leq 0.67985$ & $-0.5\leq \theta \leq 0.5$ & $-0.5\leq \psi \leq 0.5$ \\ 
  \hline
\end{tabular}
\end{center}
\caption{Normalized input constraints for input variables, $\rho$, $\theta$, $\psi$ of the ACAS Xu property $2$, $\phi_2$.}
\label{table-origRequirement}
\end{table}
\begin{table}[htbp]
\scriptsize
\begin{center}
\begin{tabular}{ | m{8em} | m{1.5cm}| m{1.5cm}| m{1.5cm}| } 
  \hline
 sub-requirement & $\rho$ bounds & $\theta$ bounds & $\psi$ bounds\\
  \hline
  1 & [0.6, 0.62] & [-0.5, -0.25] & [-0.5, -0.25] \\ 
  \hline
  2 & [0.6, 0.62] & [-0.5, -0.25] & [-0.25, 0;0] \\
  \hline
  3 & [0.6, 0.62] & [-0.5, -0.25]& [0.0, 0.25]\\ 
  \hline
  4 & [0.6, 0.62]& [-0.5, -0.25] & [0.25,  0.5]  \\
  \hline
  5 & [0.6, 0.62] & [-0.25, 0.0] &[-0.5, -0.25] \\
  \hline
  6 & [0.6, 0.62] & [-0.25, 0.0]& [-0.25, 0.0]\\
  \hline
  7 & [0.6, 0.62] & [-0.25, 0.0] & [0.25, 0.5]\\
  \hline
  8 & [0.6, 0.62] & [-0.25, 0.0] & [0.0, 0.25]\\
  \hline
  9 & [0.6, 0.62] & [0.0, 0.25] & [-0.5, -0.25]\\
  \hline
  10 & [0.6, 0.62]&[0.0, 0.25] & [-0.25, 0.0]\\
  \hline
  11 & [0.6, 0.62] & [0.0, 0.25] & [0.0, 0.25]\\
  \hline
  12 & [0.6, 0.62] & [0.0, 0.25] & [0.25, 0.5]\\
  \hline
  13 & [0.6, 0.62] & [0.25, 0.5]& [-0.5,-0.25]\\
  \hline
  14 & [0.6, 0.62] & [0.25, 0.5] & [-0.25, 0.0]\\
  \hline
  15 & [0.6, 0.62] & [0.25, 0.5] & [0.0,  0.25]\\
  \hline
  16 & [0.6, 0.62] & [0.25, 0.5] & [0.25, 0.5]\\
  \hline
  17 & [0.62, 0.64] & [-0.5, -0.25] & [-0.5, -0.25]\\
  \hline
  18 & [0.62, 0.64] & [-0.5, -0.25] & [-0.25, 0.0]\\
  \hline
  19 & [0.62, 0.64] & [-0.5,-0.25] & [0.0, 0.25]\\
  \hline
  20 & [0.62, 0.64] & [-0.5, -0.25] & [0.25, 0.5]\\
  \hline
  21 & [0.62, 0.64]& [-0.25, 0.0]& [-0.5, -0.25 ]\\
  \hline
  22 & [0.62, 0.64]& [-0.25, 0.0]& [-0.25, 0.0]\\
  \hline
  23 & [0.62, 0.64] & [-0.25, 0.0]& [0.0,  0.25]\\
  \hline
  24 & [0.62, 0.64] & [-0.25, 0.0]& [0.25, 0.5]\\
  \hline
  25 & [0.62, 0.64]& [0.0, 0.25]& [-0.5,-0.25]\\
  \hline
  26 & [0.62, 0.64] & [0.0, 0.25]& [-0.25,  0.0]\\
  \hline
  27 & [0.62, 0.64] & [0.0, 0.25] & [0.25, 0.5]\\
  \hline
  28 & [0.62, 0.64] & [0.0, 0.25] & [0.25, 0.5] \\ 
  \hline
  29 & [0.62, 0.64] & [0.25, 0.5] & [-0.5, -0.25] \\
  \hline
  30 & [0.62, 0.64] & [0.25, 0.5]& [-0.25, 0.0]\\ 
  \hline
  31 & [0.62, 0.64]& [0.25, 0.5] & [0.0, 0.25]  \\
  \hline
  32 & [0.62, 0.64] & [0.25, 0.5] & [0.25, 0.5] \\
  \hline
  33 & [0.64, 0.66] & [-0.5, -0.25]& [-0.5, -0.25]\\
  \hline
  34 & [0.64, 0.66] & [-0.5, -0.25] & [-0.25, 0.0]\\
  \hline
  35 & [0.64, 0.66] & [-0.5, -0.25] & [0.0,  0.25]\\
  \hline
  36 & [0.64, 0.66] & [ -0.5, -0.25] & [0.25, 0.5]\\
  \hline
  37 & [0.64, 0.66] & [-0.25,  0.0] & [-0.5, -0.25]\\
  \hline
  38 & [0.64, 0.66] & [-0.25,  0.0] & [-0.25, 0.0]\\
  \hline
  39 & [0.64, 0.66] & [-0.25,  0.0] & [0.0,  0.25]\\
  \hline
  40 & [0.64, 0.66] & [-0.25,  0.0]& [0.25,  0.5]\\
  \hline
\end{tabular}
\end{center}
\caption{Some of the generated sub-requirements for Property $\phi_2$ from the combination of the sub-intervals.}
\label{table-subrequirements}
\end{table}
\subsection{Implementation}
\begin{figure}[h]
\centering
\includegraphics[width=0.45\textwidth]{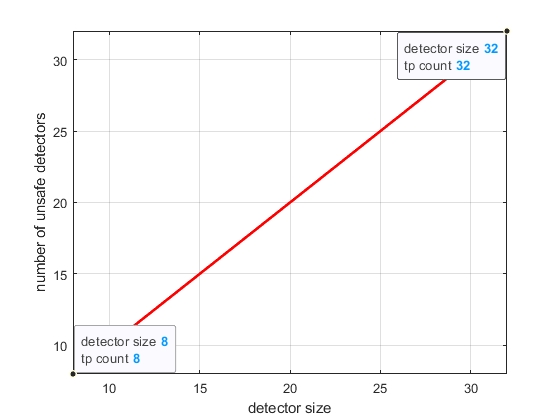}
\caption{ $tp$ count of detectors for different detector sizes using a radius value of $0.05$.}
\label{different detectors}
\end{figure}

The ground truth safe input sub-requirements are generated by verifying the sub-requirements generated in Sec.\ref{sub-requirement} using the Marabou framework \cite{katz2019marabou} on the $34$ networks with \textbf{unsat} verification outcomes. We considered the $34$ networks because they are unsafe on verifying property $2$. Among the $64$ sub-requirements, $32$ safe input sub-requirements were obtained and consequently the remaining $32$ were the unsafe input sub-requirements. Ground truth safe input sub-requirements are obtained to guide the generation of the detectors. Knowing the ground truth safe input sub-requirements, we generated varying detector sizes for the NSA procedure using a radius value between $10^{-6}$ - $0.5$. In Fig. \ref{different detectors}, we computed the true positive ($tp$) count for each detector size generated using a radius value of $0.05$ (an arbitrarily chosen radius value). $tp$ counts refers to the number of detectors that are truly unsafe after being validated by the Marabou framework. The goal is to demonstrate how effective the NSA approach is as detector sizes increase. Furthermore, we investigated the effect of the radius values on the NSA procedure, different radius values were used. We aim to show how sensitive the NSA procedure performance is with radius values within the stated bounds. It is observed that the $tp$ of the detectors size was almost constant for different radius values within the $10^{-6}$ - $0.5$ range. Table \ref{table_subrequirements} and Table \ref{table_NSA subrequirements} highlight the ground truth candidate unsafe input sub-requirements and $tp$ unsafe input sub-requirements using a detector size of $8$, respectively. 
\begin{table}[htbp]
\scriptsize
\begin{center}
\begin{tabular}{ |c|c|c|c| } 
  \hline
  $\rho$ bounds & $\theta$ bounds & $\psi$ bounds\\ 
  \hline
  [0.62, 0.64] & [0.0, 0.25] & [0.25, 0.5]\\ 
  \hline
  [0.64, 0.66] & [-0.25, 0.0] & [0.0, 0.25]\\ 
  \hline
  [0.66, 0.68] & [-0.25, 0.0]& [0.0, 0.25]\\ 
  \hline
  [0.62, 0.64] & [-0.25, 0.0] & [0.25, 0.5]\\
  \hline
  [0.6, 0.62] & [-0.25,  0.0 ]& [0.0, 0.25]\\
  \hline
  [0.66, 0.68] & [-0.25,  0.0]& [0.25, 0.5]\\
  \hline
  [0.64,  0.66] & [0.0,  0.25]& [-0.25, 0.0]\\
  \hline
  [0.62, 0.64] & [-0.25, 0.0]& [-0.25, 0.0]\\
  \hline
\end{tabular}
\end{center}
\caption{Candidate unsafe input sub-requirements using a detector size of $8$ during the NSA procedure.}
\label{table_subrequirements}
\end{table}

\begin{table}[htbp]
\begin{center}
\scriptsize
\begin{tabular}{ |c|c|c| } 
  \hline
 $\rho$ bounds & $\theta$ bounds & $\psi$ bounds\\ 
  \hline
  [0.62, 0.64] & [0.0, 0.25] & [0.25, 0.5]\\ 
  \hline
  [0.64, 0.66] & [-0.25, 0.0] & [0.0, 0.25]\\ 
  \hline
  [0.66, 0.68] & [-0.25, 0.0]& [0.0, 0.25]\\ 
  \hline
  [0.62, 0.64] & [-0.25, 0.0] & [0.25, 0.5]\\
  \hline
  [0.6, 0.62] & [-0.25,  0.0 ]& [0.0, 0.25]\\
  \hline
  [0.66, 0.68] & [-0.25,  0.0]& [0.25, 0.5]\\
  \hline
  [0.64,  0.66] & [0.0,  0.25]& [-0.25, 0.0]\\
  \hline
  [0.62, 0.64] & [-0.25, 0.0]& [-0.25, 0.0]\\
  \hline
\end{tabular}
\end{center}
\caption{ The true positive ($tp$) unsafe input sub-requirements from the $8$ detectors using the NSA procedure.}
\label{table_NSA subrequirements}
\end{table}
\subsection{Results discussion}
\begin{figure}[h]
\centering
\includegraphics[width=0.65\linewidth]{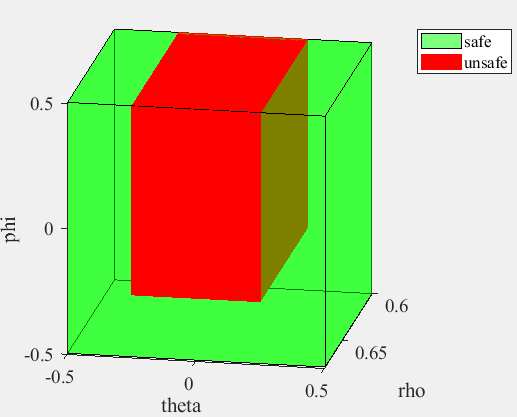}
\caption{ Safe and Unsafe regions within the generated sub-requirements. The NSA procedure is used to retrieve the Unsafe regions.}
\label{different-radius}
\end{figure}

We present the $tp$ of the NSA procedure for different detector sizes. We find for each detector size, the detectors from the partitioned input space. Table \ref{table_NSA subrequirements} presents the true positive of $8$ detectors/candidate unsafe input sub-requirements presented in Table \ref{table_subrequirements}. Out of the $8$ detectors presented, $8$ are $tp$ $\--$ thus after validating the detectors using the Marabou framework, all of them returned \textbf{sat} assignment. None of the $8$ detectors identified by the NSA procedure were false positive ($fp$). Also, we reported the $tp$ rate for varying detector sizes. Fig.\ref{different detectors} shows the $tp$ for $8$, $16$, $24$ and $32$ detectors. Since we are only evaluating the effectiveness of the NSA procedure for varying detector sizes, it is run once for each detector size. It can be observed that as the number of detector sizes increases, the number of true positives increases. For instance, given $8$ detectors, $8$ were $tp$ and with $32$ detectors, $32$ were also $tp$. Given that the partitioned input space has $32$ ground truth unsafe input sub-requirements, we evaluated the precision of the NSA procedure $5$ times for detectors size $32$.  On average $31$ were $tp$ as retrieved by the NSA procedure and can be concluded that the NSA procedure has a precision rate of $97\%$. The NSA procedure is based on a similarity measure, so we provided a range of radius values to assess the sensitivity of the NSA procedure performance. We realized empirically that, for radius values within this range, the $tp$ rate for each detector size was almost the same.Moreover, Fig.\ref{different-radius} shows a visual representation of the unsafe and safe operating regions of network within the sub-requirements; the red region depicts the unsafe regions retrieved by the NSA procedure. We can infer from the safety $\phi_2$ perspective for each of the $34$ networks when in action with the normalized input as $\rho: [0.6-0.67985]$, $\theta: [-0.25-0.25]$ and $\psi: [-0.25-0.5]$, the property is highly likely to be violated.   
\section{Conclusions and Future work}\label{Conclusion}
In this paper, a meta-heuristic approach, NSA is proposed to determine unsafe candidate regions within the bounds of a given safety requirement. A simple yet effective strategy for generating sub-intervals is proposed based on which input sub-requirements are generated. Ground truth safe regions that guides the unsafe candidate sub-requirements generations are determined from the sub-requirements using the Marabou framework. The NSA procedure effectiveness is assessed using varying detectors size and $tp$ counts for each detectors size was validated using the Marabou verification framework. Empirical results and comparison showed that NSA procedure is highly effective at retrieving candidate unsafe sub-requirements from the partitioned input space and scales well to large input sub-requirement spaces. Furthermore, we provided the range of radius values that work perfectly well for the NSA procedure. Moreover, since most verification frameworks search an entire input space to return a counterexample in cases of violation, verifying candidate unsafe input regions can significantly reduce the execution time of most verification frameworks and as well support verification-guided test case generation. However, the precision of our initial NSA procedure in determining unsafe candidate sub-requirements highly depends on the radius value setting, with radius values outside of the range provided, the $tp$ rates of the NSA procedure decreases. In the future, we will focus on using adaptive variable radius length and distributed NSA procedures which are more resilient to the curse of dimensionality, and able to apply self-learning/adaptiveness in the process of detector generation.


\section*{Acknowledgment}
The authors would like to thank the Office of the Secretary of Defense for the financial support under agreement
number FA8750-15-2-0116. This work is partially supported by the NASA University Leadership
Initiative (ULI) under grant number 80NSSC20M0161. The authors also acknowledge support from the OUSD(R\&E)/RT\&L under Cooperative Agreement Number W911NF-20-2-0261.

\bibliographystyle{IEEEtran}
\typeout{}
\bibliography{Reference}
\vspace{12pt}
\end{document}